%% file: main.tex
\documentclass[runningheads]{llncs}
\usepackage[T1]{fontenc}
\DeclareUnicodeCharacter{FF0C}{,}
\usepackage{graphicx,verbatim}
\usepackage{amsfonts}
\usepackage{amsmath}
\usepackage{amssymb}
\usepackage{tabularx}
\usepackage{adjustbox}
\usepackage{booktabs}   
\usepackage[table]{xcolor}   
\usepackage{caption}
\usepackage{multirow}
\usepackage{amssymb,pifont}
\newcommand{\cmark}{\ding{51}}

\usepackage{hyperref}
\begin{document}
%

\title{Plug-and-Play Logit Fusion for Heterogeneous Pathology Foundation Models}
%

\author{
Gexin Huang\thanks{Equal contribution.}\inst{1,3} \and
Anqi Li$^{\star}$\inst{2} \and
Yusheng Tan\inst{2} \and
Beidi Zhao\inst{1,3} \and
Gang Wang\inst{1} \and
Zu-hua Gao\inst{1} \and
Xiaoxiao Li\inst{1, 3}
}
\authorrunning{Gexin Huang et al.}
\institute{
University of British Columbia, Vancouver, BC, Canada \\
\email{gexinml@gmail.com, beidiz@student.ubc.ca, gang.wang1@bccancer.bc.ca, zuhua.gao@ubc.ca, xiaoxiao.li@ece.ubc.ca}
\and
Washington University in St. Louis, St. Louis, MO, USA \\
\email{anqi.li1@wustl.edu, t.yusheng@wustl.edu}
\and
Vector Institute, Toronto, ON, Canada
}
  
\maketitle              
\begin{abstract}
Pathology foundation models (FMs) have become central to computational histopathology, offering strong transfer performance across a wide range of diagnostic and prognostic tasks. 
The rapid proliferation of pathology foundation models creates a model-selection bottleneck: no single model is uniformly best~\cite{neidlinger2025benchmarking,ma2025pathbench}
, yet exhaustively adapting and validating many candidates for each downstream endpoint is prohibitively expensive. 
We address this challenge with a \emph{lightweight} and novel model fusion strategy, \textbf{LogitProd}, which treats independently trained FM-based predictors as fixed experts and learns sample-adaptive fusion weights over their slide-level outputs.
The fusion operates purely on logits, requiring no encoder retraining and no feature-space alignment across heterogeneous backbones. 
We further provide a theoretical analysis showing that the optimal weighted product fusion is guaranteed to perform at least as well as the best individual expert under the training objective.
We systematically evaluate LogitProd on \textbf{22} benchmarks spanning WSI-level classification, tile-level classification, gene mutation prediction, and discrete-time survival modeling. LogitProd ranks first on 20/22 tasks and improves the average performance across all tasks by $\sim$3$\%$ over the strongest single expert.
LogitProd enables practitioners to upgrade heterogeneous FM-based pipelines in a plug-and-play manner, achieving multi-expert gains with $\sim$12$\times$ lower training cost than feature-fusion alternatives.
The code is available 
\href{https://github.com/gexinh/LogitProd.git}{here}.

\keywords{Pathology \and Foundation models \and Ensemble learning.}


\end{abstract}

\input{introduction.tex}
\input{method.tex}
\input{experiments.tex}
\input{conclusion.tex}


\bibliographystyle{splncs04}
\bibliography{refs}

\clearpage
\appendix

\end{document}

%% file: introduction.tex
\section{Introduction}
\label{sec:intro}
With hematoxylin and eosin (H\&E) stained whole-slide image (WSI) digitization and deep learning, computational pathology has enabled automated diagnosis, risk stratification, and biomarker discovery from routine clinical specimens.
A standard WSI pipeline tessellates a slide into patches, encodes them into representations, and applies multiple instance learning (MIL) to aggregate patch features into slide- or patient-level predictions under weak supervision~\cite{ilse2018attention,lu2021dataefficient,shao2021transmil,xu2025milfm}.
Recently, pathology foundation models (FMs) have become strong, general-purpose encoders for these pipelines (e.g., UNI~\cite{uni2024}, CONCH~\cite{conch2024}, Virchow~\cite{virchow2024}, Prov-GigaPath~\cite{gigapath2024}), but their rapid proliferation has created a heterogeneous model zoo with substantial differences in pretraining data, architectures, and objectives.
Large-scale benchmarks increasingly suggest that no single FM is uniformly best across tasks, and that different FMs can exhibit complementary strengths even on the same endpoint~\cite{neidlinger2025benchmarking,ma2025pathbench}.
As a result, FM selection becomes a practical bottleneck: exhaustively adapting and validating many FMs for each new cohort or task is computationally expensive, while committing to a single convenient choice can leave performance unrealized.

To alleviate this model-selection bottleneck, existing solutions in computational pathology largely pursue training-time integration of multiple FMs.
First, some works intervene at the pretraining stage by distilling multiple teachers into a single, more generalizable foundation model (e.g., GPFM~\cite{ma2025generalizable}).
Second, a growing body of integration methods combines multiple pathology FMs for a downstream task by learning representation-level fusion or alignment, often through offline or online distillation to obtain task-aligned embeddings and predictors~\cite{lei2025shazam,yang2025fusion,luo2025ensemble}.
Although effective, these strategies face practical barriers at WSI scale:
(i) \emph{optimization overhead}: they entail substantial training to obtain a student model and/or a fusion-aware downstream head (e.g., 16$\times$80 GB H800 GPUs per cohort~\cite{ma2025generalizable});
(ii) \emph{re-encoding overhead}: they require re-encoding many patches or slides with multiple encoders to build fused representations, which is costly in compute, I/O, and storage; and
(iii) \emph{inflexibility}: they typically demand re-tuning the representation fusion pipeline when experts are added or cohorts shift, limiting plug-and-play upgrades.

A natural alternative is inference-time integration using only prediction outputs.
Prediction-level fusion, such as probability averaging, fixed product rules, or majority voting—provides a training-free way to combine models~\cite{dietterich2000ensemble,hinton2002training,lakshminarayanan2017simple}. Yet, in a heterogeneous FM-based expert pool, fixed fusion rules are inherently limited: expert confidence is rarely calibrated across different encoders, and an individual expert's reliability varies drastically across different slides. While learned gating (like Mixture-of-Experts) can address this by weighting experts adaptively~\cite{wu2025learning}, standard routing mechanisms rely on high-dimensional input features (e.g., patch embeddings)—the exact computational bottleneck we aim to bypass
These observations motivate a complementary question: 
\emph{Can we achieve sample-adaptive fusion of heterogeneous pathology experts from prediction outputs, without re-encoding features or retraining experts?}

To address this gap, we propose \textbf{LogitProd}, a lightweight logit-level product fusion framework for plug-and-play integration of independently trained FM-based experts.
LogitProd starts from a collection of heterogeneous experts for the same endpoint, where each expert pairs a frozen FM encoder with a task-specific head and outputs prediction logits.
Our key insight is that, even without accessing patch embeddings, expert logits contain informative \emph{reliability cues}, e.g., confidence/uncertainty statistics and inter-expert disagreement, that can guide sample-specific fusion.
Accordingly, LogitProd learns a minimal gating network to predict sample-adaptive, nonnegative expert weights directly from these logit-derived cues. Crucially, rather than using standard additive blending, LogitProd aggregates experts via a \emph{weighted product-of-experts} (PoE) formulation. This multiplicative design naturally sharpens consensus when experts agree and strictly suppresses overconfident errors from unreliable experts on a given slide.
By operating purely at the prediction level, LogitProd composes experts across encoders and embedding dimensionalities without patch re-encoding, feature alignment, or retraining a new MIL aggregator, and supports incremental expert-pool expansion by (re)fitting only a lightweight gating network on logits, with negligible additional overhead beyond running the existing experts.
We make three contributions:
(i) LogitProd, a plug-and-play logit-level product fusion method that composes independently trained pathology FM--MIL experts without patch re-encoding, feature alignment, or MIL retraining;
(ii) a theoretical analysis showing that under product fusion there exists a weighting whose risk is no worse than the best individual expert, motivating the product form; and
(iii) systematic validation on 22 benchmarks spanning WSI/tile classification, long-tailed mutation prediction, and survival, together with an accuracy--efficiency analysis against representation-fusion baselines.

%% file: method.tex
\section{Fusion of Pathology Models}
\label{sec:method}

\begin{figure}[t]
    \centering
    \includegraphics[width=0.95\linewidth]{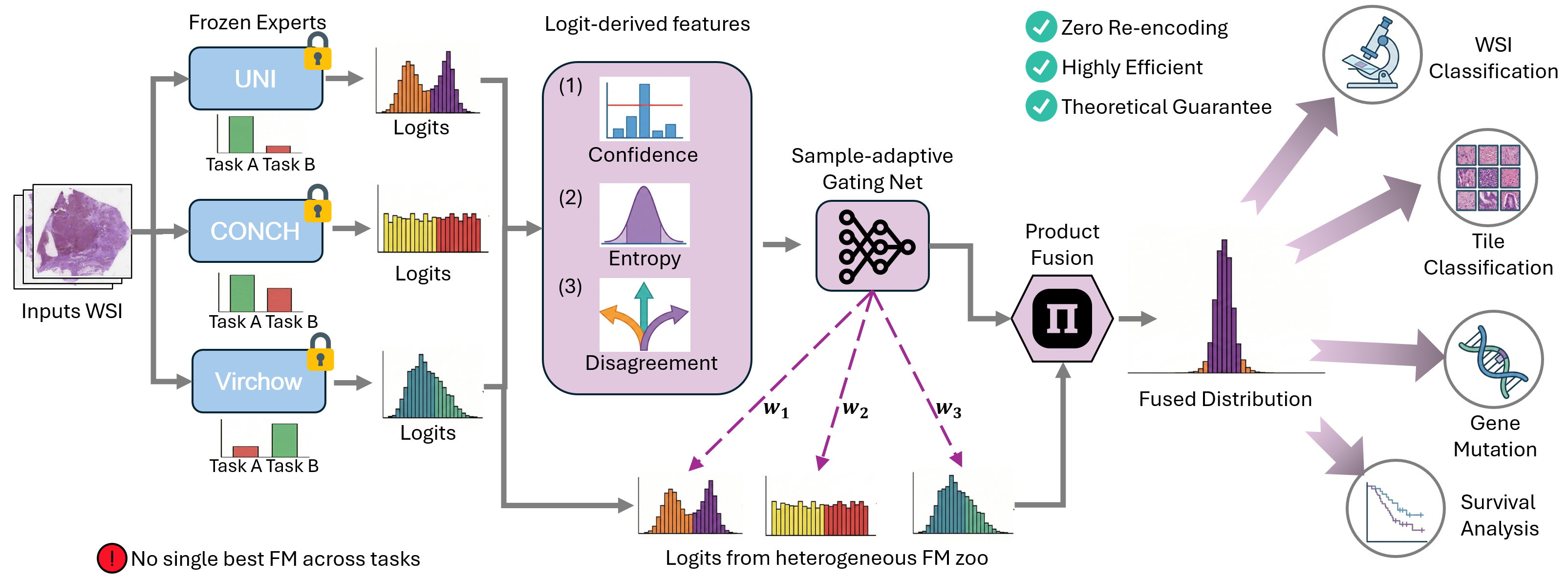}
    \caption{\textbf{Overview of LogitProd.} Frozen FM experts output logits; LogitProd derives confidence/entropy/disagreement cues to predict sample-adaptive weights and fuses experts via weighted product fusion, enabling efficient multi-task prediction without re-encoding or feature alignment.}
    \label{fig_1}
    \vspace{-0.3cm}
\end{figure}

\subsection{Problem Formulation}
We study weakly supervised learning on WSIs using multiple pretrained pathology FMs as frozen experts. 
Let $\mathcal{D}=\{(\mathcal{X}_i,y_i)\}_{i=1}^{N}$ denote a patient-level dataset, where $\mathcal{X}_i$ is the set of diagnostic WSIs for patient $i$. 
For classification, $y_i\in\{1,\dots,K\}$; for survival analysis, $y_i=(t_i,\delta_i)$ and we adopt a discrete-time formulation with $K$ time bins.

Following standard WSI preprocessing, each patient is processed by $M$ independently trained FM-based predictors $\{f_m\}_{m=1}^{M}$, each consisting of a frozen FM encoder paired with a task-specific prediction head. 
For patient $i$, expert $m$ outputs prediction logits $\mathbf{z}_{i,m}=f_m(\mathcal{X}_i)\in\mathbb{R}^{K}$, which parameterize a categorical distribution for classification or a bin-wise risk model for survival. 
We stack expert outputs as $\mathbf{Z}_i=[\mathbf{z}_{i,1}^{\top},\dots,\mathbf{z}_{i,M}^{\top}]^{\top}\in\mathbb{R}^{M\times K}$.

Our goal is to learn a logit-only fusion module that produces nonnegative expert weights $\mathbf{w}_i\in\Delta^{M-1}$ from $\mathbf{Z}_i$ and combines the $M$ predictive distributions into a fused prediction $p_{\theta}(y\mid\mathcal{X}_i)$. 
The weights are parameterized by a lightweight fusion network $g_{\theta}$ operating on logit-derived features of $\mathbf{Z}_i$ (Section~\ref{sec:method_fusion}) and trained by minimizing the task loss, using cross-entropy for classification and negative log-likelihood for survival, while keeping all experts $\{f_m\}_{m=1}^{M}$ frozen.

\subsection{LogitProd: Logit-level Product Fusion of Pathology FMs}
\label{sec:method_fusion}

Given expert logits $\mathbf{Z}_i \in \mathbb{R}^{M \times K}$ for patient $i$, LogitProd performs logit-only fusion of heterogeneous frozen predictors. 
Working purely at the prediction level avoids feature alignment and retraining, but raw logits from independently trained predictors can be scale-mismatched and highly correlated, making gating brittle and vulnerable to correlated failure modes when multiple predictors err together. 
LogitProd therefore builds compact logit-derived cues that summarize per-expert confidence and inter-expert disagreement, enabling sample-adaptive weighting that can down-weight unreliable consensus and emphasize the most reliable predictor(s) on hard cases.

\noindent\textbf{Logit-derived Gating Features.}
We first apply a standard per-expert temperature scaling \cite{guo2017calibration} to improve logit comparability across predictors: for expert $m$, a scalar $\tau_m>0$ is fitted on a held-out calibration fold by minimizing the task negative log-likelihood, and logits are calibrated as $\tilde{\mathbf{z}}_{i,m}=\mathbf{z}_{i,m}/\tau_m$. 
We then convert calibrated logits to probabilities $\mathbf{p}_{i,m}=\mathrm{softmax}(\tilde{\mathbf{z}}_{i,m})$ and extract three cues:
i) the maximum predicted probability $s_{i,m}=\max_{k} p_{i,mk}$; 
ii) the top-2 probability margin $\gamma_{i,m}=p_{i,m(k_1)}-p_{i,m(k_2)}$, where $k_1$ and $k_2$ index the largest and second-largest probabilities in $\mathbf{p}_{i,m}$; and 
iii) the predictive entropy $h_{i,m}=-\sum_{k=1}^{K} p_{i,mk}\log p_{i,mk}$.
Stacking across experts gives vectors $\mathbf{s}_i,\boldsymbol{\gamma}_i,\mathbf{h}_i\in\mathbb{R}^{M}$. 
To capture inter-expert disagreement, we compute
\begin{equation}
\bar{h}_i = \frac{1}{M}\sum_{m=1}^{M} h_{i,m}, \qquad
\bar{\mathbf{p}}_i = \frac{1}{M}\sum_{m=1}^{M}\mathbf{p}_{i,m}, \qquad
u_i = H(\bar{\mathbf{p}}_i)-\bar{h}_i,
\label{eq:disagree_feat}
\end{equation}
where $H(\bar{\mathbf{p}}_i)=-\sum_k \bar{p}_{ik}\log \bar{p}_{ik}$. 
The disagreement score $u_i$ increases when experts are individually confident yet predict different classes, having high entropy of the mean, which helps the gate detect unreliable consensus under correlated errors. 
We concatenate these cues into the gating input
\begin{equation}
\mathbf{x}_i=\mathrm{concat}\!\left(\mathbf{s}_i,\boldsymbol{\gamma}_i,\mathbf{h}_i,\bar{h}_i,u_i\right)\in\mathbb{R}^{3M+2}.
\label{eq:gating_input}
\end{equation}

\noindent\textbf{Sample-adaptive Gating.}
A lightweight gating network $g_{\theta}$ maps $\mathbf{x}_i$ to nonnegative expert weights. 
For classification, it outputs $\mathbf{w}_i=g_{\theta}(\mathbf{x}_i)\in\Delta^{M-1}$. 
For discrete-time survival with $K$ bins, it outputs $\mathbf{W}_i\in\mathbb{R}^{M\times K}$ whose columns $\mathbf{w}_i^{(k)}$ lie on the simplex. $g_{\theta}$ is a two-layer MLP with 64 hidden units and ReLU, followed by a softmax; for discrete-time survival, it uses $K$ independent gates of the same architecture (one per time bin, no parameter sharing).

\noindent\textbf{Logit-level Product Fusion.}
Given expert probabilities $\{\mathbf{p}_{i,m}\}$ and weights $\mathbf{w}_i$, LogitProd forms the fused predictive distribution via a normalized weighted product:
\begin{equation}
p_{\theta}(y \mid \mathcal{X}_i)
=
\frac{1}{Z_i}
\prod_{m=1}^{M} p_{i,m}(y)^{w_{i,m}},
\qquad
Z_i = \sum_{y'} \prod_{m=1}^{M} p_{i,m}(y')^{w_{i,m}}.
\label{eq:poe_fusion}
\end{equation}
For survival, we apply the same product fusion independently to each time bin $k$ using $\mathbf{w}_i^{(k)}$. 
The fusion module is trained using the task loss, while all expert predictors remain frozen. As shown next, for both classification and bin-wise survival there exists a choice of weights such that the product-federated model is no worse in cross-entropy risk than the best individual predictor.

\subsection{Theoretical Analysis of Logit-level Product Fusion}
\label{sec:theory}
This section provides a theory-backed justification for our central claim. 
In a strict logit-only setting where predictors are trained independently and are not jointly aligned or retrained, product fusion admits an optimal weighting whose cross-entropy risk is no worse than that of the best individual predictor.

\noindent\textbf{Proposition 1 (Classification).}
Let $\mathcal{Y}=\{1,\dots,K\}$ and let $p_{\mathrm{data}}$ be the true label distribution. 
Each predictor $m\in\{1,\dots,M\}$ induces a categorical distribution $p_m$.
Consider the product-federated distribution $p_{\mathbf{w}}$ obtained by Eq.~\eqref{eq:poe_fusion} with a fixed weight vector $\mathbf{w}\in\Delta^{M-1}$, and denote its normalization constant by $Z(\mathbf{w})$.
Then there exists $\mathbf{w}^\star\in\Delta^{M-1}$ such that
\begin{equation}
\mathcal{H}(p_{\mathrm{data}},p_{\mathbf{w}^\star})
\;\le\;
\min_{m\in\{1,\dots,M\}} \mathcal{H}(p_{\mathrm{data}},p_m),
\label{eq:theory_main}
\end{equation}
where $\mathcal{H}(p_{\mathrm{data}},q)=\mathbb{E}_{Y\sim p_{\mathrm{data}}}[-\log q(Y)]$.

\noindent\textbf{Proof Sketch.}
For each $y$, the weighted geometric mean is upper-bounded by the weighted arithmetic mean:
$\prod_{m} p_m(y)^{w_m} \le \sum_{m} w_m p_m(y)$ for $\mathbf{w}\in\Delta^{M-1}$.
Summing over $y$ yields $Z(\mathbf{w})\le 1$ and thus $\log Z(\mathbf{w})\le 0$.
Using Eq.~\eqref{eq:poe_fusion}, the cross-entropy expands as
\begin{equation}
\mathcal{H}(p_{\mathrm{data}},p_{\mathbf{w}})=
\sum_{m=1}^{M} w_m\,\mathcal{H}(p_{\mathrm{data}},p_m)+
\log Z(\mathbf{w})
\;\le\;
\sum_{m=1}^{M} w_m\,\mathcal{H}(p_{\mathrm{data}},p_m).
\label{eq:theory_decomp}
\end{equation}
Choosing $\mathbf{w}$ to be one-hot recovers the best single predictor on the right-hand side. Therefore, the minimizer $\mathbf{w}^\star$ over $\Delta^{M-1}$ satisfies Eq.~\eqref{eq:theory_main}.

\noindent\textbf{Corollary 1 (Discrete-Time Survival).}
In discrete-time survival, the negative log-likelihood decomposes into a sum of bin-wise binary cross-entropies. 
Applying Proposition~1 independently to each time bin (with bin-specific simplex weights) and summing over bins implies that there exists a collection of bin-wise weights for which the overall survival loss of product fusion is no worse than that of the best individual predictor.

Overall, our analysis shows that within the logit-level product-federated family, there exists a (globally optimal) fixed weighting whose cross-entropy risk is no worse than that of the best individual predictor. 
We view this result as a lower bound on the capacity of product fusion rather than a guarantee on the learned gate. In practice, we learn sample-adaptive weights from logit-derived cues to exploit instance-level variation, which improves numerical conditioning across heterogeneous predictors and yields more robust fusion empirically.

%% file: experiments.tex
\section{Experiments}
\label{sec:exp}


\subsection{Implementation Details}
\noindent\textbf{Setting.} We follow a standard weakly supervised WSI pipeline. Each diagnostic WSI is tiled into fixed-size patches at a predefined magnification, with background removed by tissue masking. Patches are embedded using nine pretrained pathology FMs: CONCHv1.5~\cite{conch2024}, UNI2-h~\cite{uni2024}, Phikon-v2~\cite{phikon2023}, Virchow2~\cite{virchow2024}, CTransPath--CHIEF~\cite{ctranspath2022}, H-optimus-1~\cite{hoptimus1}, Kaiko~\cite{kaiko2024}, Lunit~\cite{lunit2023}, and Prov--GigaPath~\cite{gigapath2024}. For each dataset--FM pair, embeddings are computed once and reused across tasks. 
We use 5-fold stratified cross-validation for downstream tasks. For each task, we train a task-specific predictor on frozen FM features (ABMIL for slide-level tasks and an MLP classifier for tile-level tasks), producing a pool of independently trained FM-based predictors under identical train/validation/test splits. We further reserve a small held-out subset from the training split that is not used for training or early stopping of the FM-based predictors. We fit per-expert temperature scaling on this held-out subset and train LogitProd on the resulting fixed calibrated logits using the same task loss, while keeping all FM-based predictors frozen. We select models on the validation set and report performance on the test set.

\noindent\textbf{Dataset.} We evaluate 22 benchmarks across four task families. WSI classification: TCGA-BRCA, TCGA-CRC, BRACS-3, BRACS-7, PANDA. Tile classification: CRC-100K, CCRCC, CRC-MSI, PanCancer-TIL, ESCA, PCAM. Mutation prediction: five driver genes (TP53, PIK3CA, NF1, PTEN, ARID1A) on TCGA-BRCA and TCGA-LUSC. Survival analysis: six TCGA cohorts (BRCA, CRC, BLCA, KIRC, LUSC, GBMLGG), with time-to-event discretized into $K$ bins.

\noindent\textbf{Metrics.} We report AUC/ACC/F1 for classification and C-index for survival, and summarize the accuracy--efficiency trade-off with EffScore. For classification, $\mathrm{Perf}=(\mathrm{AUC}+\mathrm{ACC}+\mathrm{F1})/3$. Cost uses FLOPs $F$, parameters $P$, and training time $H$, normalized to LogitProd ($F_0,P_0,H_0$) and combined by a geometric mean: $\mathrm{Cost}=\left(\frac{F}{F_0}\right)^{1/3}\left(\frac{P}{P_0}\right)^{1/3}\left(\frac{H}{H_0}\right)^{1/3}$. Then $\mathrm{EffScore}=\frac{\mathrm{Perf}/\mathrm{Perf}_0}{\mathrm{Cost}}$, with $\mathrm{EffScore}=1$ for LogitProd and higher values indicating better performance at lower cost.

\begin{figure*}[!t]
	\centering
	\includegraphics[width=0.95\textwidth]{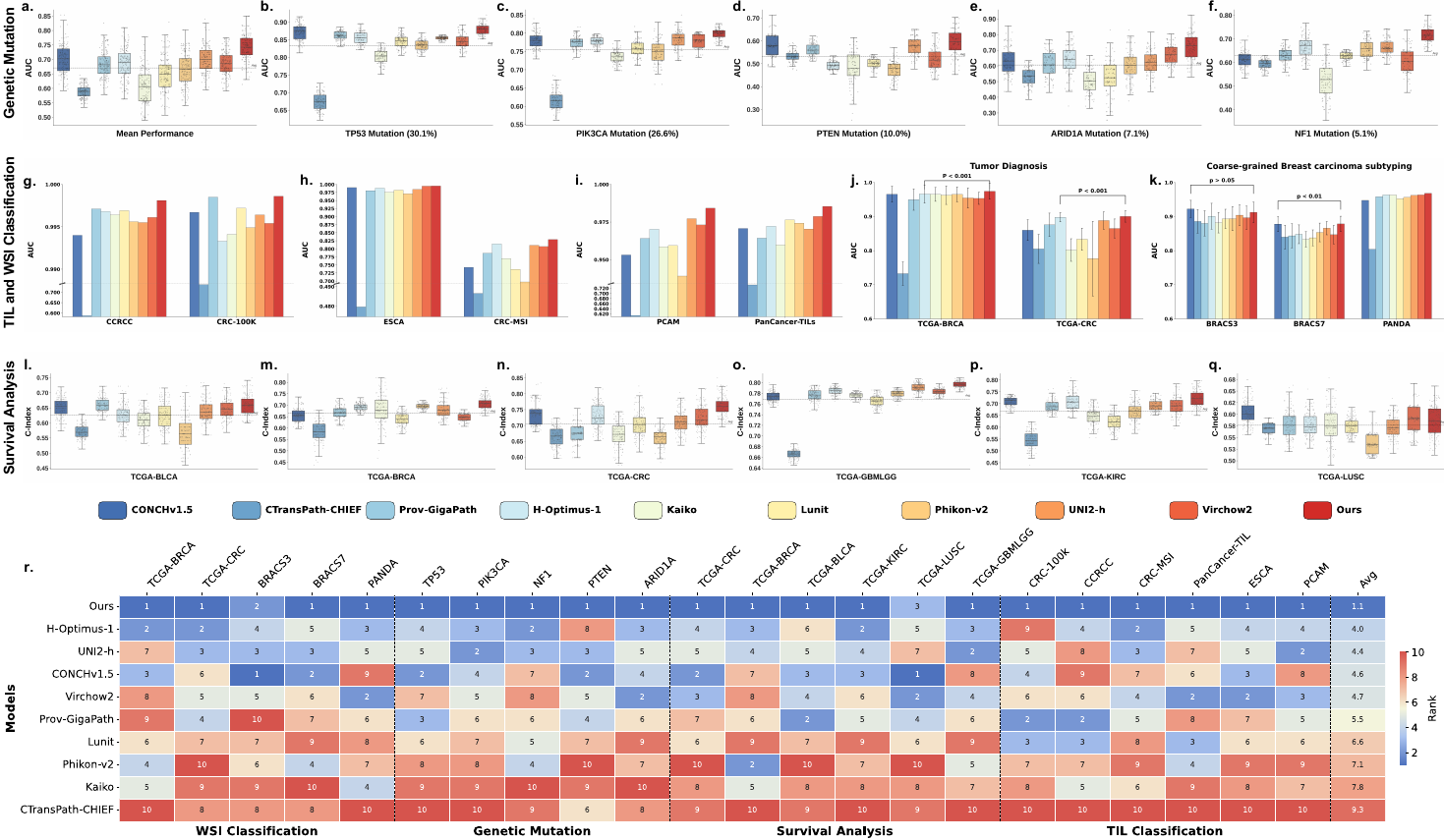}
    \caption{\textbf{Evaluation across 22 pathology tasks.}
    \textbf{a--f,} Gene mutation prediction (mAUC): \textbf{a}, mean across five genes; \textbf{b--f}, per-gene performance with prevalence.
    \textbf{g--i,} TIL classification (AUC) across six datasets.
    \textbf{j,} WSI-level tumour diagnosis (AUC).
    \textbf{k,} Breast carcinoma subtyping (AUC).
    \textbf{l--q,} C-index distributions across six TCGA cohorts for all FM-based experts and LogitProd. Box plots summarize cross-validation folds.
    \textbf{r,} Task-wide rank heatmap. The rightmost column reports mean rank, and dashed lines separate task groups.
    }
    \label{fig_2_3}
    \vspace{-0.3cm}
\end{figure*}

\subsection{Multi-task Performance Comparison and Stability}
LogitProd demonstrates consistent performance gains across diverse task families (Fig.~\ref{fig_2_3}). Specifically, it achieves the highest mean AUC in five gene mutation prediction tasks (Fig.~\ref{fig_2_3}a--f), outperforming FMs by up to 5.3\% (0.7216 vs. 0.6671 in ARID1A). For tile-level classification, LogitProd achieves the highest AUC across six benchmarks (Fig.~\ref{fig_2_3}g--i), with a notable margin in CRC-MSI (0.8288 vs. 0.8150). In WSI-level diagnosis, LogitProd improves TCGA-BRCA from 0.9658 to 0.9736 (Fig.~\ref{fig_2_3}j). It also remains competitive in subtyping (Fig.~\ref{fig_2_3}k), increasing the PANDA AUC to 0.9680. Even on BRACS-3, where gains are typically harder to obtain, LogitProd maintains a narrow margin (0.9121 vs. 0.9223). Across six TCGA cohorts for survival analysis, LogitProd consistently ranks at or near the top (Fig.~\ref{fig_2_3}l--q), with the BRCA AUC reaching 0.7338 compared to the second best 0.6975. Overall, LogitProd ranks first in 20 out of 22 tasks, achieving an average performance gain of \(\sim\)3\% over the best single expert (Fig.~\ref{fig_2_3}r).

\subsection{Efficiency Analysis}


Table~\ref{tab:fusion_efficiency_final} compares LogitProd with representative slide- and patch-level feature-fusion strategies. Feature-level baselines require retraining a new prediction head and incur substantially higher computational costs, parameters, and training time. In contrast, LogitProd performs prediction-level fusion using a lightweight module on frozen predictors. LogitProd achieves best Perf (0.9274) while maintaining high efficiency. Specifically, it utilizes fewer trainable parameters (0.77M) and achieves a $\sim$12$\times$ reduction in training time compared to patch-level fusion baselines (0.89h vs.\ 10.91h). Overall, this achieves the best performance--efficiency trade-off and demonstrates that multi-expert gains can be realized without costly feature-level fusion or model retraining.

\subsection{Ablation Study}
Table~\ref{tab:ablation_logitfed} includes standard logit-level ensemble baselines such as mean probability averaging, majority voting, and uniform product-of-experts, which helps disentangle generic ensembling gains from LogitProd-specific design choices. 
LogitProd consistently performs best against these baselines, indicating that the improvements are not explained by ensembling alone. 
Among the learnable variants, disabling sample-adaptive weighting or logit-derived features causes the largest degradation, suggesting that the gains mainly come from feature-driven, instance-wise reweighting rather than the aggregation form alone.

\input{tab_abl_eff}

%% file: tab_abl_eff.tex
\begin{table}[t]
\centering
\begin{minipage}[t]{0.48\textwidth}
\centering
\caption{\textbf{Efficiency comparison of fusion strategies}.}
\label{tab:fusion_efficiency_final}
\resizebox{\textwidth}{!}{%
\begin{tabular}{lcccccc}
\toprule
\textbf{Method} & \textbf{FLOPs (G) $\downarrow$} & \textbf{Params (M) $\downarrow$} & \textbf{Time (h) $\downarrow$} & \textbf{Perf $\uparrow$} & \textbf{EffScore $\uparrow$} \\
\midrule
\multicolumn{6}{l}{\textit{Slide-level fusion baselines}} \\
Mean Pooling& 1.90 & 1.43 & 0.90 & 0.9060 & 0.80 \\
Max Pooling  & 1.90 & 1.43 & 0.90 & 0.8897 & 0.79 \\
MLP (3-layer) & 1.91 & 13.23 & 0.91 & 0.9171 & 0.38 \\
Attention    & 1.92 & 8.99 & 0.94 & 0.9196  & 0.42 \\
\multicolumn{6}{l}{\textit{Patch-level fusion baselines}} \\
Feature Mean   & 1.44 & 1.63 & 10.91 & 0.9109 & 0.34 \\
Feature Concat & 12.19 & 12.32 & 10.63 & 0.9116 & 0.09 \\
Patch Concat   & 3.44 & 3.12 & 10.67 & 0.8903 & 0.21 \\
\midrule
\textbf{Ours} & \textbf{1.90} & \textbf{0.77} & \textbf{0.89} & \textbf{0.9274} & \textbf{1.00} \\
\bottomrule
\end{tabular}%
}
\end{minipage}
\hfill
\begin{minipage}[t]{0.48\textwidth}
\centering
\caption{\textbf{Ablation of LogitProd.} 
We ablate product aggregation (prod), sample-adaptive weighting (adap), and logit-derived features (feat).
}
\label{tab:ablation_logitfed}
\resizebox{\textwidth}{!}{
\begin{tabular}{lccc|cc|c}
\toprule
\multirow{2}{*}{\textbf{Method}} &
\multirow{2}{*}{\textbf{prod}} &
\multirow{2}{*}{\textbf{adapt}} &
\multirow{2}{*}{\textbf{feat}} &
\multicolumn{2}{c|}{\textbf{TCGA-BRCA}} &
\textbf{TCGA-CRC} \\
\cmidrule(lr){5-6}\cmidrule(lr){7-7}
& & & & \textbf{AUC} $\uparrow$ & \textbf{ACC} $\uparrow$ & \textbf{C-index} $\uparrow$ \\
\midrule
Majority vote      & -- & -- & -- & 95.2 $\pm$ 2.5 & 94.2 $\pm$ 0.9 & 59.7 $\pm$ 6.3 \\
Mean               & -- & -- & -- & 96.4 $\pm$ 2.3 & 94.4 $\pm$ 1.5 & 62.4 $\pm$ 10.4 \\
Uniform product    & \cmark & -- & -- & 96.3 $\pm$ 2.9 & 93.7 $\pm$ 1.0 & 63.3 $\pm$ 9.0 \\
Learnable sum      & -- & \cmark & \cmark & 97.1 $\pm$ 2.2 & 94.7 $\pm$ 1.8 & 73.9 $\pm$ 6.7 \\
Learnable product  & \cmark & \cmark & -- & 97.1 $\pm$ 2.3 & 94.5 $\pm$ 1.6 & 74.1 $\pm$ 10.8 \\
\rowcolor{yellow!20}
LogitProd           & \cmark & \cmark & \cmark & \textbf{97.3 $\pm$ 2.3} & \textbf{95.0 $\pm$ 1.6} & \textbf{75.8 $\pm$ 6.6} \\
\bottomrule
\end{tabular}%
}
\end{minipage}
\vspace{-0.5cm}

\end{table}

%% file: conclusion.tex
\section{Conclusion and Limitations}
\label{sec:conclusion}

We presented LogitProd, a logit-only product fusion framework for aggregating heterogeneous pathology FMs for supervised WSI analysis. 
LogitProd learns lightweight, sample-adaptive fusion weights from expert outputs and combines predictions via a weighted product, requiring neither encoder retraining nor feature-space alignment. We provide a theoretical justification that there exists a product-fusion weighting whose training objective is no worse than the best individual expert, and empirically validate LogitProd across 22 benchmarks spanning WSI/tile classification, gene mutation prediction, and discrete-time survival analysis, with a favorable performance--efficiency trade-off against feature-level fusion baselines. LogitProd currently assumes that experts are independently trained for the same endpoint and does not directly support federating predictors that target different label spaces or objectives. Because fusion operates on frozen experts, performance ultimately depends on the quality of the expert pool and cannot recover when all experts are uniformly poor. Future work includes online expert selection under distribution shift and extending logit-level fusion to multimodal foundation models.